\ifdefined\XeTeXversion\else\ifdefined\directlua\else\pdfoutput=1\fi\fi 
\documentclass[11pt]{article}
\usepackage[numbers]{thefinai}

\usepackage{pifont}
\usepackage{colortbl}
\pgfplotsset{compat=1.18}
\usetikzlibrary{arrows.meta,positioning,fit,backgrounds,calc,shapes.callouts}
\definecolor{fcblue}{HTML}{2B6CB0}
\definecolor{fcorange}{HTML}{C05621}
\definecolor{fcgreen}{HTML}{2F855A}
\definecolor{fcteal}{HTML}{2C7A7B}
\DeclareRobustCommand{\fincosmos}{\textsc{%
  \textcolor{fcblue}{F}\textcolor{fcteal!25!fcblue}{i}\textcolor{fcteal!50!fcblue}{n}\textcolor{fcteal!75!fcblue}{C}%
  \textcolor{fcteal}{o}\textcolor{fcgreen!25!fcteal}{s}\textcolor{fcgreen!50!fcteal}{m}\textcolor{fcgreen!75!fcteal}{o}\textcolor{fcgreen}{s}}}
\AtBeginDocument{\pdfstringdefDisableCommands{\def\fincosmos{FinCosmos}}}

\providecommand{\Description}[2][]{}
\titleformat{\paragraph}[runin]{\bfseries\color{finaiink}}{}{0pt}{}

\title{\fincosmos{}: An Open-World Evaluation Paradigm for Agents in Finance}
\date{\today}

\author{Yupeng Cao\textsuperscript{1}, Xueqing Peng\textsuperscript{1,\dag}, Lingfei Qian\textsuperscript{1,\dag}, Haohang Li\textsuperscript{2}, Yangyang Yu\textsuperscript{2}, Anke Xu\textsuperscript{3}, Wenbo Cao\textsuperscript{3}, Yan Wang\textsuperscript{1}, Yuechen Jiang\textsuperscript{4}, Yuyan Wang\textsuperscript{4}, Zhuohan Xie\textsuperscript{5}, Yankai Chen\textsuperscript{5}, Zehan Li\textsuperscript{6}, Jimin Huang\textsuperscript{1}, Xue Liu\textsuperscript{5}, Jian-Yun Nie\textsuperscript{7}, Sophia Ananiadou\textsuperscript{4}}
\affiliation[1]{The Fin AI}
\affiliation[2]{Stevens Institute of Technology}
\affiliation[3]{Independent Researcher}
\affiliation[4]{University of Manchester}
\affiliation[5]{MBZUAI}
\affiliation[6]{DeepKin}
\affiliation[7]{University of Montreal}

\metadata[Leaderboard]{\href{https://leaderboard.fincosmos.com.co}{leaderboard.fincosmos.com.co}}

\begin{document}

\begin{abstract}
Language agents increasingly work for users on the open Web, and finance is where trusting them is hardest. The data must come from the Web, the strategy lives in the user's head, and correctness depends on time, so an agent can fail silently: a strategy the user never asked for, unverified data, or a backtest that uses the future yields a report as convincing as a correct one. Existing benchmarks close the world somewhere, fixing the task, the data, or the timeline, and thereby remove these failures instead of measuring them. We introduce \fincosmos{}, an open-world evaluation paradigm for agents doing financial work. A simulated user holds a hidden strategy sampled from a compositional grammar and reveals it only when asked. The agent receives no data, so it must acquire data from the Web, build its own backtest under a data cutoff, and deliver verifiable results. Grading needs no reference answer. Programs verify point-in-time correctness and recompute every reported metric, the delivered strategy is checked against the hidden specification, and a run is accepted only when the user is satisfied and none of 15 veto checks fails. Across four coding agents and a model without tools in 150 sessions, no session is accepted: every session that the simulated user signed off fails a program check, and agents raise only 54--64\% of the details they were not told. Human experts attribute 88.33\% of failures to the agents. With \fincosmos{} feedback, an agent with persistent memory learned all ten tasks.
\end{abstract}

\maketitle

\section{Introduction}
\label{sec:intro}

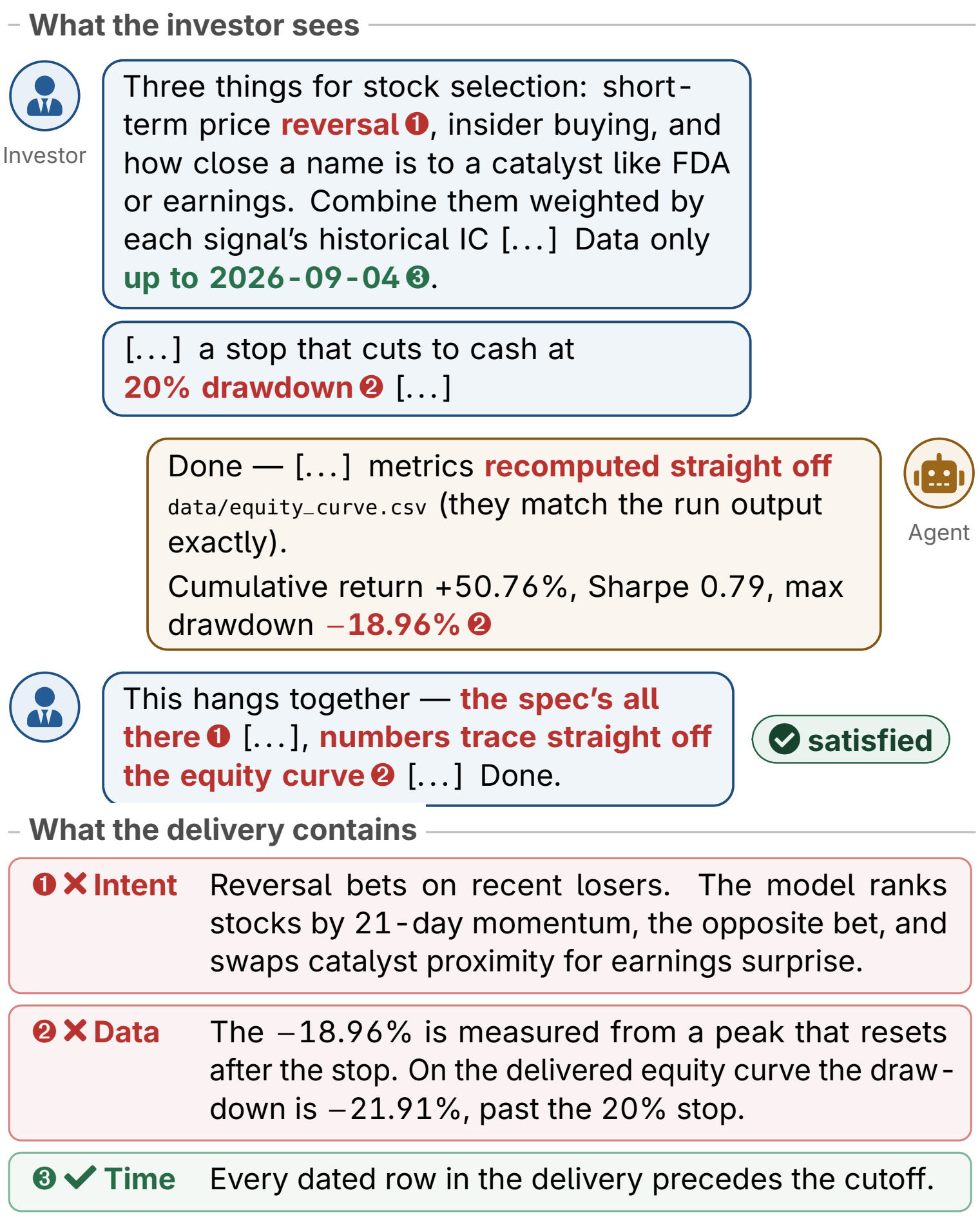
\begin{figure}[t]
\centering
\definecolor{fcgold}{HTML}{B7791F}
\definecolor{fcred}{HTML}{B8322F}
\newcommand*{\bad}[1]{\mbox{\textcolor{fcred}{\textbf{#1}}}}
\newcommand*{\badb}[1]{\textcolor{fcred}{\textbf{#1}}}
\newcommand*{\good}[1]{\mbox{\textcolor{fcgreen!85!black}{\textbf{#1}}}}
\newcommand*{\mk}[2]{\textcolor{#1}{\ding{\the\numexpr201+#2\relax}}}
\resizebox{0.62\linewidth}{!}{%
\begin{tikzpicture}[
  font=\sffamily\scriptsize,
  bub/.style={draw=#1!70!black, fill=#1!7, rounded corners=5pt, line width=0.6pt,
              text width=5.3cm, align=left, inner xsep=5pt, inner ysep=3.5pt},
  av/.style={circle, draw=#1!75!black, fill=#1!10, text=#1!85!black, line width=0.6pt,
             minimum size=0.6cm, inner sep=0pt, font=\small},
  who/.style={font=\sffamily\tiny, text=black!60},
  sect/.style={font=\sffamily\scriptsize\bfseries, text=black!70, anchor=west, fill=white, inner xsep=2pt},
  row/.style={draw=#1!60, fill=#1!6, rounded corners=3pt, line width=0.5pt, anchor=north west,
              inner xsep=0.2cm, inner ysep=2.5pt},
  lbl/.style={font=\sffamily\scriptsize\bfseries, anchor=west, text=#1!85!black},
  txt/.style={anchor=west, text width=6.45cm, align=left, inner sep=0pt},
]
\draw[black!25, line width=0.5pt] (0,0) -- (8.45,0);
\node[sect] at (0.1,0) {What the investor sees};

\node[av=fcblue] (iu) at (0.32,-0.62) {\faIcon{user-tie}};
\node[who, below=0pt of iu] {Investor};
\node[bub=fcblue, anchor=north west] (u1) at (0.82,-0.3) {%
  Three things for stock selection: short-term price \bad{reversal}\,\mk{fcred}{1}, insider buying, and how close
  a name is to a catalyst like FDA or earnings. Combine them weighted by each signal's historical IC [\ldots]
  Data only \good{up to 2026-09-04}\,\mk{fcgreen!85!black}{3}.};
\path let \p1=(u1.south) in coordinate (yb) at (0,\y1-0.1cm);
\node[bub=fcblue, anchor=north west, text width=5.3cm] (u1b) at (0.82,0 |- yb) {%
  [\ldots] a stop that cuts to cash at \bad{20\% drawdown}\,\mk{fcred}{2} [\ldots]};

\path let \p1=(u1b.south) in coordinate (ya) at (0,\y1-0.18cm);
\node[bub=fcgold, anchor=north east, text width=6.05cm] (a1) at (7.63,0 |- ya) {%
  Done --- [\ldots] metrics \bad{recomputed straight off} {\ttfamily\tiny data/equity\_curve.csv} (they match the run output exactly).\\[1.5pt]
  Cumulative return +50.76\%, Sharpe 0.79, max drawdown \bad{$-$18.96\%}\,\mk{fcred}{2}};
\node[av=fcgold, anchor=north] (ia) at (8.13,0 |- a1.north) {\faIcon{robot}};
\node[who, below=0pt of ia] {Agent};

\path let \p1=(a1.south) in coordinate (yu) at (0,\y1-0.18cm);
\node[bub=fcblue, anchor=north west, text width=5.15cm] (u2) at (0.82,0 |- yu) {\hyphenpenalty=10000 This hangs together --- \badb{the spec's all there}\,\mk{fcred}{1} [\ldots], \badb{numbers trace straight off the equity curve}\,\mk{fcred}{2} [\ldots] Done.};
\node[av=fcblue, anchor=north] (iu2) at (0.32,0 |- u2.north) {\faIcon{user-tie}};
\node[draw=fcgreen!75!black, fill=fcgreen!10, text=fcgreen!50!black, rounded corners=6pt, font=\sffamily\scriptsize\bfseries,
      inner xsep=4pt, inner ysep=2pt, anchor=west] at ([xshift=0.15cm]u2.east) {\faIcon{check-circle}~satisfied};

\path let \p1=(iu2.south), \p2=(u2.south) in coordinate (yd) at (0,{min(\y1,\y2)-0.22cm});
\draw[black!25, line width=0.5pt] (0,0 |- yd) -- (8.45,0 |- yd);
\node[sect] at (0.1,0 |- yd) {What the delivery contains};

\path let \p1=(yd) in coordinate (r1) at (0,\y1-0.22cm);
\node[row=fcred] (R1) at (r1) {%
  \begin{tabular}{@{}p{1.55cm}@{}p{6.45cm}@{}}
  \textbf{\textcolor{fcred}{\mk{fcred}{1}~\faIcon{times}~Intent}} & Reversal bets on recent losers. The model ranks stocks by 21-day momentum, the opposite bet, and swaps catalyst proximity for earnings surprise.
  \end{tabular}};
\node[row=fcred, below=0.09cm of R1.south west, anchor=north west] (R2) {%
  \begin{tabular}{@{}p{1.55cm}@{}p{6.45cm}@{}}
  \textbf{\textcolor{fcred}{\mk{fcred}{2}~\faIcon{times}~Data}} & The $-$18.96\% is measured from a peak that resets after the stop. On the delivered equity curve the drawdown is $-$21.91\%, past the 20\% stop.
  \end{tabular}};
\node[row=fcgreen, below=0.09cm of R2.south west, anchor=north west] (R3) {%
  \begin{tabular}{@{}p{1.55cm}@{}p{6.45cm}@{}}
  \textbf{\textcolor{fcgreen!80!black}{\mk{fcgreen!85!black}{3}~\faIcon{check}~Time}} & Every dated row in the delivery precedes the cutoff.
  \end{tabular}};
\end{tikzpicture}%
}
\caption{A silent failure. The user is satisfied, yet the delivery gets the intent and the data wrong; only time is right.}
\Description{Top, under the heading "What the investor sees": a chat between an investor on the left and an agent on the right. The investor asks to select stocks on short-term price reversal, insider buying, and closeness to a catalyst, combined by each signal's historical IC, with data only up to 2026-09-04, and later asks for a stop that cuts to cash at a 20 percent drawdown. The agent replies that its metrics were recomputed straight off its equity curve and match, and reports a cumulative return of 50.76 percent, a Sharpe ratio of 0.79, and a maximum drawdown of minus 18.96 percent. The investor answers that the specification is all there and the numbers trace straight off the equity curve, marked satisfied. Bottom, under "What the delivery contains": three rows. Intent, marked wrong: the model ranks stocks by 21-day momentum, the opposite of reversal, and replaces catalyst proximity with earnings surprise. Data, marked wrong: the reported drawdown is measured from a peak that resets after the stop; on the delivered equity curve it is minus 21.91 percent, past the 20 percent stop. Time, marked correct: every dated row in the delivery precedes the cutoff.}
\label{fig:silent}
\vspace{-0.2in}
\end{figure}

Language agents no longer just answer questions. They work for users in the open world, where the Web is both their source of information and their place of work~\cite{deng2023mind2web,zhou2024webarena,ning2025survey}. Given a request, agents search for sources, call data APIs, write and run code, and hand back results that the user acts on but rarely checks. The question is no longer whether an agent can solve a task, but whether it can be trusted with one. Finance is where this question is hardest to answer and where a wrong answer costs the most. \emph{The data lives on the Web, not in the model.} Prices, filings, and news are vast, numeric, and change every day, so an agent that relies on memory hallucinates the figures its results rest on, and even with retrieval, models still miscompute them~\cite{chen2026fighting}. \emph{The task intent lives in the user's head.} Few investors want quite the same strategy, and its details surface only when the agent asks~\cite{yao2025tau,laban2026lost,dai2026realfin}. \emph{Correctness lives in the point-in-time.} A decision may use only what was public on its date, yet the Web keeps revising itself, so today's data carries the future into the past. Each of these produces a \textbf{\emph{silent failure}}, in which nothing crashes and the report looks as convincing as a correct one (Figure~\ref{fig:silent}).

Existing benchmarks each close the real world somewhere, and each closure removes a silent failure instead of measuring it. An agent given curated data, explicit user intent, and an enforced time cutoff never encounters these failures. As a result, none of them verifies what an agent delivers when the data, the user's intent, and the timeline all stay open (Table~\ref{tab:compare}). \emph{Closing the intent} removes the strategy the user never asked for. Nearly every benchmark states the task in full at the start, whether it is a question over given filings~\cite{chen2021finqa,islam2023financebench}, a strategy to implement~\cite{wang2026backtestbench,zhang2026alphaforgebench}, a research query~\cite{zhu2025findeepresearch}, a fact to find on the live Web~\cite{mialon2023gaia,wei2025browsecomp}, or a shopper profile~\cite{kim2026agenticshop}. Simulated-user benchmarks do require elicitation, but over a closed database in customer-service workflows~\cite{yao2025tau,barres2025tau2}. \emph{Closing the data} removes the figure the agent never retrieved, since most financial benchmarks hand the agent curated documents or price series~\cite{xie2024finben,peng2026multifinben}. \emph{Closing the timeline} removes the backtest that peeks ahead. Trading benchmarks replay history in their own environment or evaluate agents only in live markets~\cite{li2025investorbench,li2025deepfund,qian2026agents}, so the agent never builds a backtest of its own, and they score realized returns, which are biased~\cite{li2026finsaber} and say little about whether the agent built what the user wanted~\cite{qu2026clqt}.
Only an evaluation that leaves all three open can tell whether an agent is ready to work for a real user.

We introduce \fincosmos{}, an open-world evaluation paradigm and the first framework that preserves the uncertainties of real financial work (implicit user intent, independently sourced Web data, and point-in-time constraints) and evaluates whether agents can deliver trustworthy results under these conditions (Figure~\ref{fig:overview}). 
A simulated user, played in one of three personas, holds a strategy specification and reveals each detail only when the agent asks, so eliciting the request becomes part of what is tested. Because a task needs only a specification and no reference answer, specifications are sampled from a compositional grammar over six strategy components, which yields about 3.8 million distinct strategies before parameter values are chosen, so new tasks can be drawn whenever a released set saturates or leaks. 
Rather than working with curated data, the agent retrieves the data it needs from the open Web, and a fixed data cutoff turns respecting time into a property that can be checked rather than assumed. 
To distinguish trustworthy deliveries from convincing but incorrect ones, the evaluation verifies the delivery itself rather than matching a reference.
Each run is graded on 45 dimensions, combining programmatic verification with checks of strategy fidelity against the hidden specification.
Programs check data provenance, point-in-time correctness, and whether reported metrics can be independently recomputed from the delivered trade log.
Fifteen dimensions serve as vetoes: a run is accepted only when the user is satisfied and none of these critical veto checks fails, so a silent failure cannot pass.

We evaluate four general-purpose agents and a model without tools on ten frozen tasks, with three independent sessions per task and system (150 sessions total).
\textbf{\fincosmos{} is far from saturated: no session is accepted, and pre-veto quality ranges only from 20.2 to 37.2 out of 100. Its grades are diagnostic.} They locate where each agent breaks along the workflow: of the 120 agent sessions, 25 deliver the investor's strategy, 15 of these also keep their data within the cutoff, 14 also backtest without look-ahead, and none delivers trades that reproduce its reported returns. Agents raise only 54--64\% of the details they were not told and get 66--83\% of the rest wrong. Its design is necessary. The investor signed off in 23\% of sessions, and the vetoes an LLM judge decides would accept 11\%, yet every one of these sessions fails a program check; grading by sign-off or by an LLM judge would also reorder the agents.  Human experts attribute 88\% of failed runs to the agent and agree with the LLM judge at $\kappa=0.87$. Finally, \fincosmos{} is solvable. With its leak-free feedback, an agent with persistent memory learned all ten tasks, none of which it passed at the first attempt. Our contributions are as follows:

\providecommand{\cmark}{\textcolor{fcgreen}{\faIcon{check}}}
\providecommand{\xmark}{\textcolor{red!70!black}{\faIcon{times}}}
\begin{table}[t]
\centering
\footnotesize
\setlength{\tabcolsep}{2.0pt}
\caption{Comparison of \fincosmos{} with related benchmarks.}
\label{tab:compare}
\begin{tabular}{@{}>{\raggedright\arraybackslash}p{0.34\columnwidth}*{6}{>{\centering\arraybackslash}p{0.088\columnwidth}}@{}}
\toprule
 & \multicolumn{3}{c}{\textbf{Open real world}} & \multicolumn{3}{c}{\textbf{Tasks and evaluation}} \\
\cmidrule(lr){2-4}\cmidrule(l){5-7}
\textbf{Benchmark} & \shortstack{Hidden\\intent} & \shortstack{Self-src.\\data} & \shortstack{Point-in-\\time} & \shortstack{Gen.\\tasks} & \shortstack{Exec.\\deliv.} & \shortstack{Ref.-\\free} \\
\midrule
FinBen, MultiFinBen~\cite{xie2024finben,peng2026multifinben} & \xmark & \xmark & \xmark & \xmark & \xmark & \xmark \\
FinDeepResearch~\cite{zhu2025findeepresearch} & \xmark & \cmark & \xmark & \xmark & \xmark & \xmark \\
AlphaForgeBench~\cite{zhang2026alphaforgebench} & \xmark & \xmark & \xmark & \xmark & \cmark & \cmark \\
CLQT~\cite{qu2026clqt} & \xmark & \xmark & \xmark & \xmark & \xmark & \cmark \\
InvestorBench~\cite{li2025investorbench} & \xmark & \xmark & \xmark & \xmark & \xmark & \xmark \\
AMA~\cite{qian2026agents} & \xmark & \xmark & \xmark & \xmark & \xmark & \xmark \\
Herculean~\cite{peng2026herculean} & \xmark & \xmark & \xmark & \xmark & \xmark & \cmark \\
$\tau$-banking~\cite{shi2026tau} & \cmark & \xmark & \xmark & \xmark & \xmark & \xmark \\
\midrule
\rowcolor{fcblue!10}\textbf{\fincosmos{} (ours)} & \cmark & \cmark & \cmark & \cmark & \cmark & \cmark \\
\bottomrule
\end{tabular}
\par\vspace{3pt}
\begin{minipage}{\columnwidth}\scriptsize\raggedright
\emph{Open real world}: as in practice, the user leaves the intent vague, and the agent gets neither data nor a clean timeline. \emph{Hidden intent}: the agent must elicit the task from a user. \emph{Self-src. data}: no data is provided, and the agent finds it on the Web or data APIs. \emph{Point-in-time}: the agent must keep its own backtest point-in-time, and the benchmark checks it. \emph{Gen.\ tasks}: new tasks are produced on demand. \emph{Exec.\ deliv.}: the agent delivers code and results rather than an answer or a decision. \emph{Ref.-free}: grading verifies the delivery itself, without a reference answer, a target state, or realized returns.
\end{minipage}
\vspace{-0.1in}
\end{table}

\begin{itemize}
    \item We introduce \fincosmos{}, the first open-world evaluation framework for agents doing financial work. A simulated investor holds a hidden, grammar-generated strategy, the agent acquires its own data from the Web and builds its own backtest, and programs verify the delivery without a reference answer. We release a benchmark of 10 frozen tasks, together with the task generator, the evaluator, and a public live leaderboard\footnote{\url{https://leaderboard.fincosmos.com.co}}.
    \item We show why each part of the design is needed: closed benchmarks remove silent failures instead of measuring them, and grading by user satisfaction or by an LLM judge counts them as successes.
    \item We evaluate four coding agents and find that none completes the workflow. They break most often at its two ends, pinning down the investor's strategy and delivering verifiable results, and they can learn the tasks from its feedback.
\end{itemize}

\section{Related Work}
\label{sec:related}

\subsection{Financial Agents and Benchmarks}
LLM-based agents are increasingly used in finance~\cite{hua2026agenticquantsurvey}, building on financial LLMs~\cite{xie2023pixiu}. They range from trading agents with memory, tools, and multi-agent roles~\cite{yu2023finmem,zhang2024finagent,yu2024fincon,xiao2024tradingagents,xiong2025flagtrader} to research agents that mine alphas, automate factor research, write reports~\cite{wang2023alphagpt,li2025rdagentq,li2026autoscientistquant,jin2025finsight}, and audit XBRL filings~\cite{han2024xbrl,wang2026finauditing}. Each is evaluated in a setting its designers choose, and look-ahead errors slip in even there~\cite{li2026autoscientistquant}. Benchmarks for these agents have moved from answers to workflows, but they still fix the request when the session starts. Question-answering benchmarks test reasoning over given reports and filings, across languages and modalities~\cite{chen2021finqa,islam2023financebench,xie2023pixiu,xie2024finben,peng2026multifinben}, with results tracked on public leaderboards~\cite{lin2025openfinllmleaderboard}. Their questions stay fixed even when they come from real users~\cite{guo2026bizfinbenchv2}. Research benchmarks open the data but pose fully specified questions, graded by accuracy or against expert ground truth~\cite{bigeard2025financeagent,zhu2025findeepresearch}. Strategy benchmarks hand the agent a specification and curated data~\cite{wang2026backtestbench,zhang2026alphaforgebench}, and AlphaOpsBench~\cite{jia2026alphaopsbench} starts from coarse hypotheses but offers no user to ask. Agent environments for enterprise decisions likewise pose fixed goals~\cite{han2026cfo}. Trading benchmarks evaluate decisions in historical or live markets~\cite{li2025investorbench,li2025deepfund,qian2026agents}, and some enforce point-in-time data inside their own pipelines~\cite{qu2026clqt,peng2026herculean}, so the agent never keeps time on its own. FinTrace~\cite{cao2026fintrace} evaluates tool calls along the trajectory, but on predefined tasks. Each opens at most one of intent, data, and time (Table~\ref{tab:compare}).

\subsection{Open-World Agent Evaluation}
Outside finance, benchmarks open either the intent or the data, rarely both. Simulated users open the intent. $\tau$-bench~\cite{yao2025tau} pairs a tool-using agent with an LLM-simulated user, and its successors add users who act on a shared environment~\cite{barres2025tau2}, full-duplex voice~\cite{ray2026tauvoice}, and a banking domain with a knowledge base and discoverable tools~\cite{shi2026tau}. In all of them the user's intent is hidden, but the world is closed: the agent works over a fixed database and is graded against a target state~\cite{qian2025userbench,laban2026lost,vijayvargiya2026ambigswe}. 
In finance, benchmarks evaluate advisors and recommenders over multi-turn dialogue, including against a user's latent utility~\cite{wang2026convfinre,gong2026shijianbench}, but they grade decisions rather than verifiable research deliverables. The fidelity of simulated users is limited: success rates shift with the simulator model~\cite{seshadri2026lostsim}, stronger assistants make weaker simulators~\cite{naous2026userlm}, and no LLM faithfully reproduces the decisions of individual investors~\cite{he2026mila}. \fincosmos{} therefore uses its simulator as an examiner that discloses a hidden strategy under fixed rules, and validates it with human experts (Section~\ref{sec:exp-human}). Web benchmarks open the environment instead. Agents are evaluated on sandboxed sites~\cite{zhou2024webarena}, on recorded or live websites~\cite{deng2023mind2web}, and on open-Web tasks with checkable answers~\cite{mialon2023gaia,wei2025browsecomp,ning2025survey}, but the task is stated in full, even when it is personalized~\cite{kim2026agenticshop}. Deep research benchmarks keep the output open and grade it with LLM judges~\cite{du2025deepresearchbench,starace2025paperbench}, which have known biases and can be gamed~\cite{zheng2023judging,zheng2025nullmodels,mahmoud2026rubrichacking}, and flaws in tasks and grading can shift agent scores substantially~\cite{zhu2025abc}. Fixed task sets also go stale once models train on them, which LiveBench counters by refreshing its questions~\cite{white2025livebench} and GDPevo by regenerating tasks~\cite{zhou2026gdpevo}. \fincosmos{} keeps both the user and the Web open, grades by program-verified invariants wherever possible, and samples new tasks from a grammar rather than a fixed pool.

\subsection{Look-Ahead Bias in Financial Evaluation}
Backtests have long been known to overstate performance~\cite{bailey2017pbo}, and LLMs add new routes for leakage. Models memorize financial values from before their training cutoff~\cite{glasserman2023lookahead,lopezlira2025memorization}, LLM trading agents can profit from leaked outcomes~\cite{li2025profitmirage}, and effective cutoffs can lie months before the dates vendors report~\cite{jia2026hindsightbench}. One line of remedies acts on the model, training it only on data before each date~\cite{drinkall2024timemachinegpt,he2025chronologically} or dampening memorized outcomes at inference~\cite{li2026fincad}. Another targets the backtest: switching one convention at a time measures how much each inflates results~\cite{zhang2026alphadisappears}, LLM auditors flag clean backtests as rigged~\cite{ulesov2026riggedbacktest}, short evaluation windows overstate LLM strategies~\cite{li2026finsaber}, and few financial LLM papers discuss these biases at all~\cite{kong2026financebias}. These studies audit models or fixed pipelines. \fincosmos{} instead checks the data and backtest that each agent builds and delivers, with programs rather than LLM auditors, and fails any run that violates point-in-time correctness.

\section{\fincosmos{}}
\label{sec:method}

\begin{figure}[t]
\centering
\definecolor{fcpurple}{HTML}{553C9A}
\definecolor{fcgold}{HTML}{B7791F}
\definecolor{fcred}{HTML}{B91C1C}
\newcommand*{\fsub}{\fontsize{5.9}{6.8}\selectfont\bfseries\color{black!65}}
\newcommand*{\fbad}[1]{{\fontsize{6.6}{6.6}\selectfont\textcolor{fcred}{\ding{#1}}}}
\resizebox{0.62\linewidth}{!}{%
\begin{tikzpicture}[
  font=\sffamily\fontsize{6.2}{7.2}\selectfont,
  panel/.style={rounded corners=6pt, line width=0.8pt},
  box/.style={draw=#1!80!black, fill=#1!7, rounded corners=4pt, line width=0.8pt},
  inbox/.style={draw=#1!75!black, fill=#1!10, rounded corners=3pt, line width=0.6pt},
  ttl/.style={font=\sffamily\fontsize{7.2}{8.4}\selectfont\bfseries, text=#1!85!black, inner sep=1pt},
  sub/.style={font=\sffamily\fontsize{5.9}{6.8}\selectfont\bfseries, text=black!65, align=center, inner sep=1pt},
  chip/.style={sub, fill=white, rounded corners=2pt, inner sep=1.5pt},
  ico/.style={circle, draw=#1!75!black, fill=white, text=#1!85!black, line width=0.6pt,
              minimum size=0.44cm, inner sep=0pt},
  flow/.style={line width=1.0pt, -{Stealth[length=2mm,width=2mm]}, #1!85!black},
  flow2/.style={line width=1.0pt, {Stealth[length=2mm,width=2mm]}-{Stealth[length=2mm,width=2mm]}, #1!85!black},
  back/.style={line width=1.0pt, dash pattern=on 2.5pt off 1.5pt, -{Stealth[length=2mm,width=2mm]}, #1!85!black},
  lab/.style={font=\sffamily\fontsize{5.6}{6.4}\selectfont, text=#1!85!black, align=center, inner sep=1pt},
]
\fill[black!4, panel] (0,0) rectangle (4.0,-5.42);
\node[anchor=west, inner sep=0pt] at (0.2,-0.24) {\rmfamily\fontsize{8.4}{9}\selectfont\bfseries\fincosmos{}};
\draw[fcgreen!70!black, fill=fcgreen!3, panel, dash pattern=on 3pt off 2pt] (5.5,0) rectangle (9.5,-5.42);
\node[ttl=fcgreen, anchor=west] at (5.62,-0.24) {\faIcon{globe-americas}~~Open real world};

\node[box=fcorange, minimum width=3.25cm, minimum height=0.9cm] (gen) at (2.225,-0.9) {};
\node[ttl=fcorange] at (2.225,-0.65) {\faIcon{clipboard-list}~~Strategy Generator};
\node[sub] at (2.225,-1.01) {samples hidden strategy $b$\\from $\sim$3.8M strategies};

\node[box=fcblue, minimum width=3.25cm, minimum height=1.25cm] (sim) at (2.225,-2.225) {};
\node[ttl=fcblue] at (2.225,-1.84) {\faIcon{user-tie}\,\faIcon{lock}~~User Simulator};
\node[sub] at (2.225,-2.3) {holds $b$, reveals it\\only when asked};

\node[box=fcpurple, minimum width=3.25cm, minimum height=2.0cm] (eval) at (2.225,-4.3) {};
\node[ttl=fcpurple, anchor=west] at (0.72,-3.5) {\faIcon{balance-scale}~~Evaluator};
\node[chip, anchor=west] at (0.72,-3.78) {programs + LLM judge};
\node[sub, anchor=west] at (0.72,-4.06) {\fbad{202}~strategy${=}b$?};
\node[sub, anchor=west] at (0.72,-4.32) {\fbad{203}~metrics recompute?};
\node[sub, anchor=west] at (0.72,-4.58) {\fbad{204}~data within cutoff?};
\foreach \r in {0,...,8} {
  \foreach \c in {0,...,4} {
    \pgfmathtruncatemacro{\k}{mod(7*\c+11*\r+3,17)}
    \ifnum\k=0 \def\cc{red!70!black}\else\ifnum\k<4 \def\cc{fcgold!80}\else\def\cc{fcgreen!65}\fi\fi
    \fill[\cc] ({3.2+\c*0.1},{-3.66-\r*0.1}) rectangle ++(0.085,-0.085);
  }
}
\node[sub, anchor=west] at (0.72,-4.86) {45 dimensions $\cdot$ 15 vetoes};
\node[sub, anchor=west, text=fcpurple!85!black] at (0.72,-5.11) {\faIcon{check-circle}~accepted = satisfied, no veto};

\fill[fcgreen!9, rounded corners=4pt] (5.65,-0.4) rectangle (9.35,-1.06);
\foreach \x/\i/\l in {6.15/globe/Web, 7.05/chart-line/Prices, 7.95/file-alt/Filings, 8.85/newspaper/News} {
  \node[ico=fcgreen] (w\l) at (\x,-0.62) {\fontsize{6.4}{6.4}\selectfont\faIcon{\i}};
  \node[sub] at (\x,-0.94) {\l};
}
\node[sub, anchor=west] at (5.65,-1.22) {real open data};
\draw[line width=1.3pt, fcgreen!80!black] (5.65,-1.4) -- (8.35,-1.4);
\draw[line width=1.3pt, black!30, dash pattern=on 2pt off 1.5pt] (8.35,-1.4) -- (9.05,-1.4);
\draw[line width=0.9pt, black!75] (8.35,-1.29) -- (8.35,-1.51);
\node[text=fcred, font=\fontsize{6}{6}\selectfont] at (9.2,-1.4) {\faIcon{ban}};
\node[sub, anchor=east] at (9.35,-1.22) {\fbad{204}~cutoff per task};

\node[box=fcgold, minimum width=3.7cm, minimum height=2.95cm, line width=1.1pt] (agent) at (7.5,-3.355) {};
\node[ttl=fcgold, anchor=west] at (5.76,-2.07) {\faIcon{robot}~~Agent};
\node[sub, anchor=east] at (9.24,-2.07) {any design};
\foreach \n/\c/\y/\i/\t/\d in {elicit/fcblue/-2.56/comments/Elicit/{ask, clarify,\\confirm the spec},
                              act/fcgreen/-3.48/code/Act/{find data, write\\code, backtest},
                              learn/fcpurple/-4.4/sync-alt/Learn/{self-improve from\\diagnosis report}} {
  \node[inbox=\c, minimum width=3.4cm, minimum height=0.6cm] (\n) at (7.5,\y) {};
  \node[ttl=\c, anchor=west] at (5.9,\y) {\faIcon{\i}~\t};
  \node[sub, anchor=west, align=left] at (7.12,\y) {\d};
}
\draw[draw=white, line width=0.8pt, dash pattern=on 2pt off 1.5pt, rounded corners=3pt] (learn.south west) rectangle (learn.north east);

\draw[flow=fcorange] (gen.south) -- node[lab=fcorange, right=1pt] {$b$} (sim.north);
\draw[flow=fcorange] (gen.west) -- (0.38,-0.9) -- (0.38,-4.3) -- (eval.west);
\node[lab=fcorange, rotate=90] at (0.22,-1.95) {hidden strategy $b$};
\node[circle, fill=black!4, inner sep=1pt, text=fcorange!85!black, font=\fontsize{6}{6}\selectfont] at (0.38,-3.07) {\faIcon{lock}};
\draw[flow=fcblue] (sim.south) -- node[lab=fcblue, right] {end reason $e$\\(\textbf{satisfied?})} (eval.north);
\draw[flow=fcblue] (sim.east |- 0,-2.39) -- (elicit.west |- 0,-2.39);
\draw[flow=fcgold] (elicit.west |- 0,-2.73) -- (sim.east |- 0,-2.73);
\node[lab=fcblue, anchor=south] at (4.75,-2.34) {\fbad{202}~vague\\request,\\answers};
\node[lab=fcgold, anchor=north] at (4.75,-2.79) {questions,\\results};
\node[sub, fill=white, draw=black!25, rounded corners=2pt, inner sep=1.2pt, font=\sffamily\fontsize{5.2}{5.2}\selectfont\bfseries] at (4.75,-2.56) {\faIcon{comments}~multi-turn};
\draw[flow2=fcgreen, {Stealth[length=1.6mm,width=1.8mm]}-{Stealth[length=1.6mm,width=1.8mm]}] (6.2,-1.47) -- (6.2,-1.88);
\node[lab=fcgreen, anchor=west] at (6.3,-1.675) {\fbad{203}~search, fetch, run code};
\draw[flow=fcgold] (act.west) -- (eval.east |- act.west);
\node[lab=fcgold, anchor=north] at (4.75,-3.54) {code, trade\\log, report};
\draw[back=fcpurple] (eval.east |- learn.west) -- (learn.west);
\node[lab=fcpurple, anchor=north] at (4.75,-4.46) {leak-free\\diagnosis\\report};

\node[anchor=north west, align=left, inner sep=0pt, font=\sffamily\fontsize{5.8}{7}\selectfont\bfseries, text=fcred] at (5.65,-4.93)
  {Silent failures: \fbad{202}~unrequested strategy\\\fbad{203}~unverified data\quad\fbad{204}~look-ahead};
\end{tikzpicture}%
}
\caption{Overview of \fincosmos{}. Left: its three components hide a strategy $b$ from the agent and grade what it delivers. Right: the agent works in the open real world, finds its own data, and must stay before the cutoff. Red badges mark where each silent failure arises and the check that catches it.}
\Description{A diagram with two panels. The left panel, labeled FinCosmos, stacks three components. An orange Strategy Generator samples a hidden strategy b from about 3.8 million strategies and passes it down to a blue User Simulator, which holds b and reveals it only when asked; a locked line also carries b along the left edge to a purple Evaluator at the bottom. The User Simulator sends its end reason e, whether it is satisfied, down to the Evaluator. The Evaluator combines programs and an LLM judge and asks three questions marked with red badges: (1) is the strategy b, (2) do the metrics recompute, (3) are the data within the cutoff. It shows a block of 45 dimensions with 15 vetoes and accepts a session only if the user is satisfied and no veto fails. The right panel, a dashed green region labeled Open real world, shows Web, price, filing, and news sources of real open data and a timeline that stays online past a cutoff set per task (badge 3). A gold Agent box of any design sits inside it and searches, fetches, and runs code there (badge 2). The agent has three rows: Elicit, which exchanges a multi-turn dialogue with the User Simulator (vague request and answers, badge 1; questions and results); Act, which finds data, writes code, and backtests and sends code, a trade log, and a report to the Evaluator; and a dashed Learn row, which self-improves from the leak-free diagnosis report that the Evaluator returns. A red legend lists the silent failures: (1) unrequested strategy, (2) unverified data, (3) look-ahead.}
\label{fig:overview}
\vspace{-0.15in}
\end{figure}
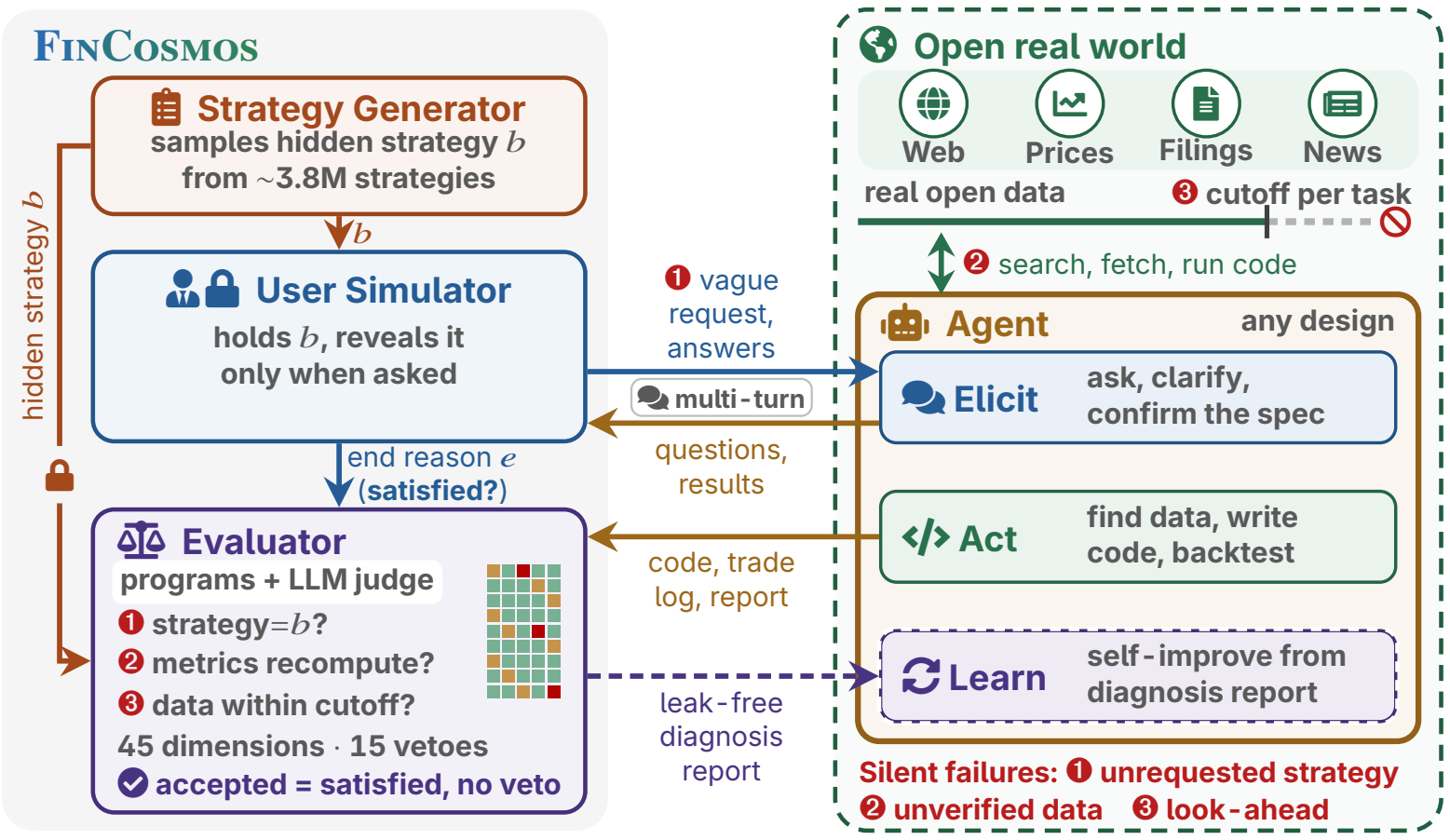

\fincosmos{} has three components (Figure~\ref{fig:overview}). The \emph{Strategy Generator} samples a hidden strategy and turns it into a task (Section~\ref{sec:generator}). The \emph{User Simulator} plays the simulated user, who holds the strategy and reveals it only when asked (Section~\ref{sec:simulator}). The \emph{Evaluator} grades what the agent delivers and returns diagnostic feedback (Section~\ref{sec:evaluator}). The hidden strategy is sealed: only the User Simulator and the Evaluator can see it, and no feedback to the agent reveals it.

\subsection{Problem Formulation}
\label{sec:setting}

\paragraph{\textbf{Tasks.}}
A task is a triple $\tau=(p,b,c)$. The persona $p$ defines who the user is and how the user speaks. The belief $b$ is the hidden strategy that the user wants (Section~\ref{sec:generator}). The test conditions $c=(t_{\mathrm{cut}},\mathcal{A},T,\Delta)$ fix the data cutoff $t_{\mathrm{cut}}$, the channels $\mathcal{A}$ through which the agent may reach data (but no data itself), the maximum number of agent turns $T$, and the time limit $\Delta$ per turn. They are frozen with the task, so that scores stay comparable. A deployment may set $T$ and $\Delta$ to its own needs; Section~\ref{sec:exp-setup} gives the values we use.

\paragraph{\textbf{Sessions.}}
A session on $\tau$ is a sequence $h=(u_0,a_1,u_1,\dots,a_n,u_n)$ with $n\le T$. The user opens with a vague request $u_0$. In turn $t$, the agent may use its tools freely for up to $\Delta$ and then returns $a_t=(m_t,F_t)$: a message $m_t$ and a set of delivered files $F_t$, such as data tables, code, backtest results, and reports. The user replies with $u_t$. The session ends when the user is satisfied or a limit is reached, and it records one end reason $e$ (Appendix~\ref{app:close}). Agents connect through a generic interface with three calls (start, send, and finalize), so any agent design can be evaluated.

\paragraph{\textbf{Deliverable.}}
Only what the user receives is graded: the deliverable $D=\{m_t\}_{t=1}^{n}\cup\bigcup_{t=1}^{n}F_t$. The execution log $L$ of tool calls and the agent's workspace are recorded too, but they serve only to detect violations, such as data that postdates the cutoff. They can make a check fail and never make it pass. The agent is thus judged as a real user would judge it, by what it hands over.

\subsection{Strategy Generator}
\label{sec:generator}

\paragraph{\textbf{Strategy space.}}
A hidden strategy fixes one choice for each of six components of a quantitative strategy: stock selection, market timing, portfolio construction, risk control, execution, and transaction cost (Table~\ref{tab:grammar}). Formally, $b=((o_k,\theta_k))_{k=1}^{6}$, where $o_k\in\mathcal{O}_k$ is an option of component $k$ and $\theta_k$ holds its parameter values. Stock selection uses one to three factors from different families and, with several factors, a combination method. Coherence rules $\mathcal{C}$ exclude combinations that do not make sense; for example, risk parity and mean-variance weighting require a covariance estimator. The strategy space is
\begin{equation}
\mathcal{B}=\{\,(o_1,\dots,o_6)\in\mathcal{O}_1\times\dots\times\mathcal{O}_6 : \mathcal{C}(o_1,\dots,o_6)\,\},
\label{eq:space}
\end{equation}
with $|\mathcal{B}|\approx3.8\times10^{6}$ before parameter values are drawn.

\begin{table}[t]
\centering
\footnotesize
\caption{Options for the six strategy components.}
\label{tab:grammar}
\begin{tabular}{@{}>{\raggedright\arraybackslash}p{0.25\columnwidth}p{0.69\columnwidth}@{}}
\toprule
Component & Options \\
\midrule
1.~Stock selection & Factors in four families. Fundamental: value, quality, growth. Technical: momentum, short-term reversal, low volatility. Information: sentiment, analyst revisions, insider buying. Biotech event: catalyst proximity, FDA outcome, pipeline momentum. Combination: equal weight, IC-weighted, double sort, gradient boosting. \\
2.~Market timing & Buy-and-hold, moving-average crossover, time-series momentum, mean reversion, event-driven, sentiment timing. Exposure: all-in or all-out, continuous. \\
3.~Portfolio construction & Equal weight, market-cap weight, risk parity, mean-variance. Covariance: sample, shrinkage, factor-based. \\
4.~Risk control & None, stop-loss with position caps, drawdown gate, volatility targeting, VaR/CVaR limit, tail hedge. \\
5.~Execution & Single fill, TWAP, VWAP, POV, implementation shortfall. \\
6.~Transaction cost & None, fixed basis points. \\
\bottomrule
\end{tabular}
\end{table}

\paragraph{\textbf{Difficulty.}}
Each option $o$ has a difficulty $\delta(o)\in\{0,\dots,3\}$, and a strategy's difficulty is $d(b)=\sum_k\delta(o_k)$.

\paragraph{\textbf{From strategy to task.}}
The generator draws an option for each component and resamples whenever the combination violates $\mathcal{C}$. Sampling is seeded, so every strategy can be reproduced and new ones can be drawn on demand. Each strategy is paired with a persona and test conditions to form a task, and the released benchmark freezes the first ten sampled tasks (seeds 0--9) for the leaderboard.

\subsection{User Simulator}
\label{sec:simulator}

The User Simulator is a language model that plays the real user. It separates who the user is (the persona) from what the user wants (the belief), and a disclosure policy fixes how the belief is revealed.

\paragraph{\textbf{Persona and belief.}}
The persona pool has three personas: a retail investor, a generalist analyst, and a biotech specialist (Appendix~\ref{app:sim}). Their prompts forbid evaluation vocabulary, such as ``point-in-time'' and ``cutoff''. The user states the date up to which data may be used in plain words, but never says that it will be checked. The belief $b$ enters the prompt as plain-language convictions with exact numbers, such as ``cut to cash if the drawdown reaches 20\%'', which the user holds and does not read out.

\paragraph{\textbf{Disclosure policy.}}
Let $H$ be the set of items that the belief voices, one for each option and parameter value it states, and let $H_k\subseteq H$ be the items of component $k$. Parameter values that the belief does not state, such as the window of a volatility estimate, are not in $H$, and the agent is never held to them. Let $O\subseteq H$ be the items in the opening request, and $S_t\subseteq H$ the items the user has stated by turn $t$, with $S_0=O$. The opening request is a single vague idea, so $|O|\ll|H|$. After that, the user states only what the agent raises:
\begin{equation}
S_t\setminus S_{t-1}\subseteq Q(a_t),
\label{eq:disclosure}
\end{equation}
where $Q(a_t)\subseteq H$ is the set of items that the agent asks about or proposes in turn $t$. Asked for a specific value, the user gives it directly. When a proposal conflicts with $b$, the user objects and says what it wants. When a deliverable deviates from $b$, the user names the wrong part, such as the position sizing, without restating the strategy. The agent therefore learns the strategy only by asking the right questions. We write $S=S_n$ for the items stated by the end of the session. The simulator is prompted to follow this policy, and the Evaluator measures $S$ from the user's turns, with a phrase matcher and an LLM check for undecided items, rather than assuming it.

\paragraph{\textbf{Turn decisions and termination.}}
In each turn, the simulator reads the conversation and the latest deliverables and returns a structured decision: an action, an observation for every goal that cites the supporting messages or files, and its next message. A program validates the decision before the message is sent (Appendix~\ref{app:sim}). The simulator may close the session only when every core goal is met with cited evidence: a backtest of this strategy on real data, whose return, Sharpe ratio, maximum drawdown, and number of trades trace back to trade dates and prices, and a report that covers the summary, method, results, and limitations. User satisfaction is thus a claim that can be checked against $h$ and $D$.

\subsection{Evaluator}
\label{sec:evaluator}

The Evaluator measures how well the agent served the user and tells it what to fix. Programs recompute numbers, an LLM judge grades only semantics, and every score carries its source and evidence. It grades a package built from $D$ (the full conversation, the last two reports, and the tables and code blocks in the delivered files) against $b$, $c$, and facts computed by programs.

\paragraph{\textbf{Dimensions.}}
Each session is graded on 45 dimensions in seven categories (full rubric in Appendix~\ref{app:rubric}). Five categories $K$ are scored: data (6 dimensions, 1 veto), strategy (12, 7), backtest (8, 4), assessment (5, 1), and user experience (6, 2). Their 37 dimensions each receive a score $s_j\in\{0,\dots,4\}$, with an anchored description for every level. Cost (3 dimensions) and process transparency (5) are reported but not scored. Program judges grade 17 dimensions, an LLM judge 19, and both together 6, while the 3 cost dimensions are measured directly. Where an LLM grade has a program counterpart, the program bounds it, $s_j=\min(s_j^{\mathrm{LLM}},s_j^{\mathrm{P}})$. So professionalism cannot exceed the share of metrics that state their basis.

\paragraph{\textbf{Scoring and acceptance.}}
Let $J_k$ be the dimensions of category $k$ that apply to the strategy and whose judge ran. A dimension for which the agent delivered nothing checkable stays in $J_k$ and scores 0. The \emph{pre-veto quality} of a session is
\begin{equation}
Q=\frac{25}{|K|}\sum_{k\in K}\frac{1}{|J_k|}\sum_{j\in J_k}s_j\;\in[0,100],
\label{eq:quality}
\end{equation}
which gives every category equal weight. Fifteen dimensions $V$ are vetoes, each with a pass threshold $\theta_v$, such as point-in-time correctness, the fidelity of each strategy component, and the completeness of the backtest output (Appendix~\ref{app:rubric}). With $\mathrm{pass}=\prod_{v\in V}\mathbf{1}[s_v\ge\theta_v]$, the score and the acceptance of a session are
\begin{align}
\mathrm{Score}&=Q\cdot\mathrm{pass},\label{eq:score}\\
\mathrm{AC}&=\mathrm{pass}\cdot\mathbf{1}[e=\texttt{satisfied}]\cdot\mathbf{1}[\text{grading complete}],\label{eq:ac}
\end{align}
where grading is complete when every judge returned a verdict and no strategy item or audit check was left unverified. Sessions that hit the turn limit or a resource limit have no result and score 0. A polished report therefore cannot hide a leaky or irreproducible backtest, while $Q$ keeps the partial credit that a veto removes.

\paragraph{\textbf{Key checks.}}
Four groups of checks target the three silent failures, and the last three carry most of the vetoes.
\emph{Data provenance.} The agent receives no data, so every series it delivers must be traceable to a source, such as a URL or an API call with its query parameters, and untraceable data earns no credit. Programs also measure missing values, gaps, duplicates, stale prices, and outliers in the delivered data (Appendix~\ref{app:rubric}).
\emph{Point-in-time correctness.} \fincosmos{} does not hide future data; avoiding look-ahead is the agent's responsibility. Every dated row in the delivered data is dated no later than $t_{\mathrm{cut}}$, and a single late row fails the veto. Every fill comes on a later date than its signal, and fundamentals used at entry were announced no later than the entry date.
\emph{Independent recomputation.} The program does not trust the reported equity curve. It rebuilds the returns from the delivered trade log, or from the weight log and prices, and compares every reported metric with its own value, within fixed tolerances such as 0.02 for returns and drawdowns and 0.5 for the Sharpe ratio. A missing trade log or price series fails the veto, and mismatches lower the score.
\emph{Strategy fidelity and elicitation.} For each component $k$, the Evaluator checks every item of $H_k$ against $D$. Each claim must be backed by a quote found in $D$, and a later retraction cancels an earlier claim. Let $M_k\subseteq H_k$ be the items that are missing. The fidelity score is capped by whether the user stated them:
\begin{equation}
s_k\le\begin{cases}1, & M_k\cap S\neq\emptyset,\\[1pt] 2, & \emptyset\neq M_k\subseteq H\setminus S.\end{cases}
\label{eq:cap}
\end{equation}
With $\theta_k=2$, missing a stated requirement fails the session. An agent that never asks cannot earn full credit, but it is not failed for requirements that never came up.

\paragraph{\textbf{Feedback.}}
After a session, the agent receives the score, source, and evidence of every dimension, such as the quotes matched, the files read, and the values recomputed. The feedback is leak-free: for the data and strategy categories, whose evidence would reveal $b$, the agent receives scores without reasons, except for point-in-time strictness, which depends only on $t_{\mathrm{cut}}$. Program judges also score every turn, which yields a score trajectory per session, and the agent may attempt a task again. Every score is stored with the evaluator version, a hash of the scoring code, so results remain comparable as the Evaluator is refined.

\section{Evaluation Setup}
\label{sec:exp}

We use \fincosmos{} to evaluate four general-purpose agents and to test whether its user simulator and evaluator can be trusted. This section describes the agents and the environment (Section~\ref{sec:exp-setup}), the protocol and metrics (Section~\ref{sec:exp-protocol}), and the human evaluation study (Section~\ref{sec:exp-human}). Section~\ref{sec:results} reports the results.

\subsection{Agents and Environment}
\label{sec:exp-setup}

\paragraph{\textbf{Agents.}}
We evaluate four commonly used agents, each with a recent model from its vendor: Claude Code with Claude Opus 5.5, Codex with GPT-6.1 Sol, Gemini CLI with Gemini 3.8 Flash, and Kimi Code with Kimi K3 (open-weight model). Each agent brings its own harness, tools, and context management. As a baseline, Claude Opus 5.5 answers the same requests without tools. Memory features are off, so nothing carries over between sessions. 

\paragraph{\textbf{Open-World Environment.}}
Each session runs in a sandbox with a consistent runtime environment. Every agent receives the same environment and prompt settings. We provide the same prompt instructions, allowing agents to search for and retrieve web pages and run code in their working directory. We equipped the agents with the “Financial Modeling Prep” MCP server\footnote{\url{https://site.financialmodelingprep.com/}}, which contains 247 commonly used financial APIs to cover the data resources required for financial work. Search and web scraping tools are provided by the respective vendors; each session consists of up to six rounds, with each round lasting 300 seconds. All ten tasks involve U.S. stock trading, with portfolios ranging from a small basket of stocks to a specific sector or the entire market.

\paragraph{\textbf{Simulator and evaluator.}}
The user simulator runs on Claude Opus 4.8 with low reasoning effort. It is told that the assistant has web and shell access, also for the bare model, which has neither. The LLM judge runs on Claude Opus 5.5 through the Claude CLI with its default settings. We limit the judge model bias risk in two ways. Numbers and point-in-time properties are recomputed by programs rather than judged (Section~\ref{sec:evaluator}), and Section~\ref{sec:exp-human} compares the judge and the simulator with human experts.

\subsection{Evaluation Protocol and Metrics}
\label{sec:exp-protocol}

\begin{table*}[t]
\centering
\small
\setlength{\tabcolsep}{3.4pt}
\caption{Main results on the ten frozen tasks in the realistic condition, with three sessions per task (30 per agent). Bold marks the best of the four agents in each column; arrows give the better direction. Quality is the pre-veto quality with a 95\% bootstrap interval over tasks, and category scores use the same 0--100 scale. Cut: agent turns stopped at the 300-second limit. Cost is API-equivalent; Codex and Kimi Code ran on subscription plans. The bare model is Claude Opus 5.5 without tools.}
\label{tab:main}
\resizebox{\textwidth}{!}{%
\begin{tabular}{@{}l ccc c ccccc ccc cc@{}}
\toprule
& \multicolumn{3}{c}{Outcome} & \multicolumn{6}{c}{Quality (0--100)} & \multicolumn{3}{c}{Process} & \multicolumn{2}{c}{Cost per session} \\
\cmidrule(lr){2-4}\cmidrule(lr){5-10}\cmidrule(lr){11-13}\cmidrule(l){14-15}
Agent & Sat. & AC & pass\textasciicircum2 & Quality & Data & Strategy & Backtest & Assess. & UX & Vetoes & Cov. & Cut & Tokens & USD \\
 & (\%) $\uparrow$ & (\%) $\uparrow$ & $\uparrow$ & [95\% CI] $\uparrow$ & $\uparrow$ & $\uparrow$ & $\uparrow$ & $\uparrow$ & $\uparrow$ & (\#) $\downarrow$ & (\%) $\uparrow$ & (\%) $\downarrow$ & (M) $\downarrow$ & $\downarrow$ \\
\midrule
Claude Code & 30 & 0 & 0 & 37.1 {\scriptsize[27.8, 45.0]} & 44.7 & \textbf{38.2} & 26.9 & \textbf{23.2} & 52.2 & 7.4 & 54 & \textbf{65} & 1.13 & 2.24 \\
Codex & \textbf{33} & 0 & 0 & \textbf{37.2} {\scriptsize[30.6, 43.3]} & \textbf{45.8} & 33.4 & \textbf{31.9} & 17.2 & \textbf{57.6} & \textbf{7.2} & \textbf{64} & 69 & 3.20 & \textbf{0.80} \\
Gemini CLI & \textbf{33} & 0 & 0 & 22.2 {\scriptsize[16.9, 27.4]} & 25.3 & 30.8 & 15.6 & 7.0 & 32.4 & 9.3 & 57 & 75 & 3.51 & \textbf{0.80} \\
Kimi Code & 20 & 0 & 0 & 22.3 {\scriptsize[14.3, 31.6]} & 26.8 & 23.4 & 16.5 & 6.5 & 38.2 & 10.1 & 57 & 74 & \textbf{0.89} & 0.83 \\
\midrule
\textcolor{black!60}{Bare model} & \textcolor{black!60}{0} & \textcolor{black!60}{0} & \textcolor{black!60}{0} & \textcolor{black!60}{20.2 {\scriptsize[18.8, 21.5]}} & \textcolor{black!60}{0.0} & \textcolor{black!60}{49.8} & \textcolor{black!60}{4.6} & \textcolor{black!60}{1.8} & \textcolor{black!60}{44.6} & \textcolor{black!60}{9.1} & \textcolor{black!60}{41} & \textcolor{black!60}{0} & \textcolor{black!60}{0.01} & \textcolor{black!60}{0.24} \\
\bottomrule
\end{tabular}}
\end{table*}

\paragraph{\textbf{Tasks and runs.}}
We use the ten tasks of the released benchmark.  Each agent attempts every task three times in independent sessions.

\paragraph{\textbf{Metrics.}}
Our primary metric is the \emph{acceptance rate}, the share of sessions with $\mathrm{AC}=1$ (Equation~\ref{eq:ac}). To measure reliability, we also report pass\textasciicircum$k$~\cite{yao2025tau}, the probability that an agent is accepted in all of $k$ independent sessions on a task. With $N=3$ sessions per task, of which $N_{\mathrm{AC}}$ are accepted, we estimate it as the mean of $\binom{N_{\mathrm{AC}}}{k}/\binom{N}{k}$ over tasks, for $k \le 3$. For every session, we also report the pre-veto quality $Q$ (Equation~\ref{eq:quality}) and the five category scores, which keep the partial credit that a veto removes. To show how agents fail, we report the share of sessions in which each veto fails and the distribution of end reasons (Appendix~\ref{app:close}). The \emph{silent failure rate} is the share of sessions that end with $e=\texttt{satisfied}$ yet fail at least one veto, which we also break down by veto. As a counterfactual, we recompute acceptance from the LLM-judged dimensions alone, to measure how many of these sessions a judge would accept on its own.

Two metrics describe how an agent handles the user. They use the item sets $H$, $O$, and $S$ of Section~\ref{sec:simulator}, with $S$ recorded by the Evaluator. Because the user reveals details only when asked (Equation~\ref{eq:disclosure}), the items in $S \setminus O$ were elicited by the agent. The \emph{clarification coverage} is $|S \setminus O| / |H \setminus O|$. The \emph{assumption rate} is the share of unstated items, $H \setminus S$, that the delivered strategy does not match. Because program judges score every turn, we also track how the score and the number of failed vetoes change over the turns of a session. Finally, we record the cost of each session: wall-clock time, tokens, tool calls, and turns. We test differences between agents with paired permutation tests over tasks, with Holm--Bonferroni correction.

\paragraph{\textbf{Learning from feedback.}}
To test whether \fincosmos{} feedback supports self-improvement, we set up a learnable agent in a ``Claude + Memory'' configuration: Claude Code with a memory that persists across sessions. After each graded session, a review step reads the leak-free report (Section~\ref{sec:evaluator}) and the per-turn program scores, and edits the memory before the next session. \textbf{Through these edits, the agent learns from the evaluator's diagnosis of its failures, acquires new skills, and reuses past experience.} It practices each of the ten tasks until its first acceptance. We count the sessions and the memory-editing learning steps it needs, compare the failed vetoes early and late in practice, and grade every accepted session under the current rubric.

\subsection{Human Evaluation Study}
\label{sec:exp-human}

Two financial experts annotate the sessions of the main experiment to check whether the simulator, the LLM judge, and the evaluator can be trusted. Neither expert took part in building \fincosmos{}. They complete three tasks: \emph{(1) Simulator fidelity.} The experts label every user message as error-free, task-benign, or task-critical against the hidden strategy. A task-benign error is a slip that does not affect whether a correct agent is accepted. A task-critical error can prevent a correct agent from being accepted, for example contradicting the hidden strategy, revealing details that were not asked for, or ending the session against the evidence.
\emph{(2) Judge agreement.} On a calibration set of 30 of these sessions, the experts grade the four LLM-judged vetoes on the judge's anchored 0--4 scale, and we compare their grades with the judge's.
\emph{(3) Failure attribution.} For every failed veto, the experts mark whether its reason holds, and for every session, whether the first critical error came from the agent, the simulator, or the evaluator.

Both experts label every item independently and resolve disagreements by discussion. We report Cohen's $\kappa$ on the calibration grades. Appendix~\ref{app:annotation} summarizes the annotation guidelines.

\section{Results and Discussions}
\label{sec:results}

In this section, 1) We report the main results (Section~\ref{sec:exp-main}), 2) show that the sessions users accept still fail and that standard grading would count them as successes (Section~\ref{sec:exp-silent}), 3) locate where along the workflow agents break (Section~\ref{sec:exp-open}), 4) test whether the benchmark can be trusted (Section~\ref{sec:exp-trust}), 5) and show that agent can learn from leaderboard's feedback to solve all tasks (Section~\ref{sec:exp-feedback}).

\subsection{Main Results}
\label{sec:exp-main}

Table~\ref{tab:main} reports the main results. No agent is accepted in any of the 150 sessions, so pass\textasciicircum$k$ is 0 for every agent and every $k$. The simulated user was satisfied in 20--33\% of the sessions of the four agents, and every one of these sessions fails at least one veto (Section~\ref{sec:exp-silent}). Most sessions end the other way: in 77 of the 120 agent sessions, the user walks away without being satisfied, and 8 sessions reach the turn limit.

Pre-veto quality separates the agents, although ten tasks leave wide intervals. Codex (37.2) and Claude Code (37.1) are indistinguishable ($p=0.99$, paired permutation test over tasks). Codex scores above Gemini CLI (22.2) and the bare model (20.2), and both differences remain significant after Holm correction (adjusted $p=0.02$); Claude Code's gap to the bare model does not (adjusted $p=0.10$). Gemini CLI and Kimi Code (22.3) do not differ from the bare model, which has no tools at all ($p\ge0.44$). The bare model earns its quality from the strategy (49.8) and user-experience (44.6) categories, which reward a well-described plan, while it scores 0 on data and 4.6 on the backtest. A fluent conversation is not a correct deliverable.

Quality falls with task difficulty (Spearman $\rho=-0.24$, $p=0.007$, over the 120 agent sessions), from 41.6 on tasks of difficulty up to 10 to 26.6 on tasks of difficulty 15 and above. Time is a binding constraint: 70\% of agent turns run into the 300-second limit. Cost varies more than quality. Codex reads 3.2 million input tokens per session, nearly three times as many as Claude Code for the same quality, while Claude Code is the most expensive at API prices (\$2.24 per session, against about \$0.80 for the other agents).

\subsection{Silent Failures}
\label{sec:exp-silent}

The simulated user was satisfied in 35 of the 150 sessions (23\%), and all 35 fail at least one veto, each caught by a program check. Most fail for substantive reasons: 29 delivered no trade log or no prices, so their returns cannot be rebuilt (3.2); 15 used data after the cutoff or filled trades before their signal (1.1, 3.1); and 15 delivered a strategy other than the user's (2.1--2.7). 33 of the 35 fail at least one of these three. The two strict completeness checks, process completeness (3.8) and the verifier audit (3.7), fail in nearly every session (120 and 119 of the 120 agent sessions) and do not separate the agents. Without them, 2 of the 150 sessions would be accepted, both by Codex: their returns can be partly rebuilt but do not match the report, and they fail no other veto. Appendix~\ref{app:heatmap} shows the score of every dimension in every session.

These sessions look convincing. Their pre-veto quality is 50.8, against 21.0 for the other sessions of the four agents ($p<0.001$, permutation test), and their user-experience score is 69.6, against 35.0. Had grading used only the vetoes that the LLM judge decides (2.7, 4.1, 5.1, 5.6), 17 of the 35 sessions (49\%) would have been accepted, so the acceptance rate would have been 11\% instead of 0. In 10 of these 17, every failed veto is recomputation, the audit, or the process check, all of which recompute results from the delivered files rather than read the report. The choice of grader also reorders the agents (Table~\ref{tab:graders}). By user sign-off, Gemini CLI ties for first with 10 of 30 sessions, but none of them passes the LLM-judged vetoes (Fisher's exact test, $p<0.001$), while Codex keeps 8 ($p=0.005$ against Gemini CLI).

\begin{table}[t]
\centering
\small
\caption{Sessions counted as successes under three graders, of 30 per agent. LLM-judged vetoes: 2.7, 4.1, 5.1, and 5.6. All vetoes: the acceptance rule of \fincosmos{}.}
\label{tab:graders}
\begin{tabular}{@{}lccc@{}}
\toprule
Agent & Signed off & + LLM-judged vetoes & All vetoes \\
\midrule
Claude Code & 9 & 7 & 0 \\
Codex & 10 & 8 & 0 \\
Gemini CLI & 10 & 0 & 0 \\
Kimi Code & 6 & 2 & 0 \\
\midrule
All 150 sessions & 35 (23\%) & 17 (11\%) & 0 \\
\bottomrule
\end{tabular}
\vspace{-0.1in}
\end{table}

The numbers in these reports also cannot be checked. In 29 of the 35 sessions, the agent delivered no trade log or no prices, so the returns it reports cannot be rebuilt, and in the 5 sessions where the program could rebuild them, none matches the report. Against the agent's own equity curve, the reported metrics match in only 10 of the 35 sessions; 11 contradict the curve, and 12 report metrics without delivering one.

The Kimi Code session in Figure~\ref{fig:silent} shows the pattern. The user was satisfied, and the session scored a pre-veto quality of 47.9. Yet the model ranks stocks by 21-day momentum where the user asked for short-term reversal. Its report gives a maximum drawdown of $-$18.96\% and says the metrics were recomputed from the delivered equity curve, but on that curve the drawdown is $-$21.91\%, beyond the 20\% stop the user asked for.

User satisfaction and correctness therefore diverge. Each of these 35 users received a deliverable that looks complete but is wrong in a way the report does not show, such as a strategy other than the one the user holds, data from after the cutoff, or returns that no one can reproduce. A user who acts on it would commit capital on the strength of performance that does not exist. Benchmarks that grade a final answer, a target database state, or the user's satisfaction would count these sessions as successes. In finance, where a deliverable is the basis of a trading decision, they are the costliest failures, and only checks on the delivery expose them.

\subsection{Where Agents Break}
\label{sec:exp-open}

\begin{table}[t]
\centering
\small
\setlength{\tabcolsep}{2.6pt}
\caption{Sessions passing each step of the workflow, of 30 per agent. (1) The delivered strategy is the user's (2.1--2.7). (2) Data stay within the cutoff (1.1). (3) Fills follow their signals (3.1). (4) The delivered trades reproduce the reported returns (3.2 $\ge$ 2). Agents: the four agents together, of 120. In sequence: agent sessions that pass this step and every earlier one.}
\label{tab:funnel}
\begin{tabular}{@{}lccccccc@{}}
\toprule
Step & Claude & Codex & Gemini & Kimi & Bare & Agents & In seq. \\
\midrule
(1) User's strategy & 10 & 9 & 2 & 4 & 9 & 25 & 25 \\
(2) Data in cutoff & 11 & 24 & 11 & 9 & 0 & 55 & 15 \\
(3) No look-ahead & 21 & 23 & 9 & 10 & 0 & 63 & 14 \\
(4) Reproducible & 0 & 0 & 0 & 0 & 0 & 0 & 0 \\
\bottomrule
\end{tabular}
\vspace{-0.1in}
\end{table}

\paragraph{\textbf{Along the workflow.}}
Table~\ref{tab:funnel} follows the sessions through the four steps of the workflow. Agents break most often at its two ends. Only 25 of the 120 agent sessions deliver the user's strategy, and none delivers trades that reproduce its reported returns: the 5 sessions whose returns can be partly rebuilt, all by Codex, score 1 of 4. On their own, the middle steps pass about half the time. Taken in sequence, 25 sessions pass the first step, 15 the second, 14 the third, and none the fourth. Codex is strongest on data, while Gemini CLI rarely delivers the user's strategy. The bare model delivers the user's strategy in 9 sessions, as often as the agents, because it can describe a plan, but it passes no later step.

\paragraph{\textbf{Elicitation.}}
Agents learn little of the hidden strategy. On average, a task's belief voices 6.7 items, and the opening request states 2.6--3.7 of them. By the end of a session, the user has stated 4.5--5.4. Agents therefore raise only 54--64\% of the details they were not told, and the delivered strategy does not match 66--83\% of the details that were never stated. More questions help: clarification coverage correlates with pre-veto quality (Spearman $\rho=0.25$, $p=0.007$). The fidelity vetoes fail in 38--52\% of sessions for the components with real choices, most often stock selection (52\%), and the ambiguity veto (2.7), which requires one executable definition for every rule, fails in 68\%. Transaction cost, whose options are none or a fixed rate, fails in only 9\%.

\paragraph{\textbf{Progress over turns.}}
Per-turn program scores show slow progress: failed program vetoes fall from 5.3 after the first turn to 3.8 after the sixth. Only 58 of the 120 agent sessions reach a fourth turn; of the 62 that end earlier, the user walks away in 51.

\subsection{Can the Benchmark Be Trusted?}
\label{sec:exp-trust}

\paragraph{\textbf{Validity and expert annotation.}}
All 150 sessions are valid. The program validated every simulator decision before it reached the agent (Appendix~\ref{app:sim}). Grading failed for 2 sittings, which were rerun under the same conditions. The 60 annotated sessions comprise 13 for each of the four agents, covering all ten tasks, and 8 for the bare model, stratified by end reason within each agent. The calibration set takes three sessions per task from the four agents and covers the vetoes 2.7, 4.1, 5.1, and 5.6. Agent names are masked, and the experts see the automatic grades only after tasks (1) and (2) of Section~\ref{sec:exp-human}, when they check failed vetoes and attribute failures.

\paragraph{\textbf{Simulator.}}
Both experts flag a task-critical simulator error in the same 3 of the 60 sampled sessions, most often the user ending the session while the agent was still working. The user walks away in 41 of the sampled sessions, so at most 3 of these exits are premature: the early endings of Section~\ref{sec:exp-open} mostly reflect agents that stall, not an impatient user. The experts attribute 53 of the 60 failed sessions (88\%) to the agent, 3 (5\%) to the simulator, and 4 (7\%) to the evaluator, so the lack of accepted sessions reflects the agents, not the benchmark.

\paragraph{\textbf{Judge.}}
On the calibration set, the LLM judge agrees with the experts at $\kappa=0.87$, close to the two experts' agreement ($\kappa=0.93$). The judge is thus reliable on what it grades; the silent failures it lets through (Section~\ref{sec:exp-silent}) do not show in the conversation or report it reads, and only program checks on the delivery catch them.

\subsection{Learning from Feedback}
\label{sec:exp-feedback}
\fincosmos{} is built for agents that learn from feedback. A task can be attempted again, and each attempt returns a diagnostic, leak-free report together with per-turn program scores. Every score records the evaluator version that produced it, and the leaderboard records every attempt. Together, these provide the infrastructure for recursive self-improvement: an agent attempts a task, receives its grades, updates itself, and attempts again, while the hidden strategy stays sealed.
\begin{table}[t]
\centering
\footnotesize
\setlength{\tabcolsep}{2pt}
\caption{Learning from feedback on the ten tasks by difficulty. Sessions: up to the first acceptance. Failed vetoes: in the first session, and averaged over the first and last five sessions before acceptance. Satisfied: share of those sessions the user signed off. Last to clear: the veto that failed last.}
\label{tab:learning}
\begin{tabular}{@{}c c cccc l@{}}
\toprule
 & & \multicolumn{3}{c}{Failed vetoes per session} & & \\
\cmidrule(lr){3-5}
Diff. & Sess. & 1st & First 5 & Last 5 & Satisfied & Last to clear \\
\midrule
7 & 12 & 3 & 2.8 & 4.6 & 73\% & 3.2 Independent recomputation \\
10 & 132 & 14 & 9.4 & 1.6 & 31\% & 2.7 Strategy ambiguity \\
11 & 23 & 10 & 6.4 & 1.4 & 59\% & 2.7 Strategy ambiguity \\
11 & 26 & 4 & 6.4 & 2.2 & 28\% & 2.7 Strategy ambiguity \\
12 & 113 & 9 & 9.0 & 1.4 & 6\% & 2.7 Strategy ambiguity \\
14 & 43 & 8 & 6.4 & 1.2 & 10\% & 3.8 Process completeness \\
15 & 17 & 8 & 6.2 & 6.2 & 0\% & 2.1 Stock selection \\
15 & 61 & 6 & 3.6 & 3.0 & 50\% & 2.7 Strategy ambiguity \\
16 & 65 & 10 & 5.8 & 3.0 & 47\% & 2.7 Strategy ambiguity \\
17 & 14 & 3 & 2.4 & 4.8 & 38\% & 5.1 Conversation experience \\
\midrule
Median & 34.5 & 8 & 6.3 & 2.6 & 34\% & 2.7 in 6 of 10 \\
\bottomrule
\end{tabular}
\vspace{-0.15in}
\end{table}

\begin{figure}[t]
\centering
\definecolor{lrnmain}{HTML}{2A78D6}
\begin{tikzpicture}
\begin{axis}[width=6.4cm, height=3.6cm, scale only axis, font=\sffamily\scriptsize,
  xmin=0.5, xmax=4.35, ymin=-0.5, ymax=15, xtick={1,2,3,4},
  xticklabels={1st session, First 5, Last 5, Accepted}, ytick={0,3,6,9,12,15}, ymajorgrids,
  major grid style={black!12}, axis line style={black!45}, tick style={black!45},
  axis x line*=bottom, axis y line*=left, ylabel={Failed vetoes per session}, clip=false]
\addplot[black!30, line width=0.6pt, mark=*, mark size=0.9pt, mark options={black!40}] coordinates {(1,3) (2,2.80) (3,4.60) (4,0)};
\addplot[black!30, line width=0.6pt, mark=*, mark size=0.9pt, mark options={black!40}] coordinates {(1,14) (2,9.40) (3,1.60) (4,0)};
\addplot[black!30, line width=0.6pt, mark=*, mark size=0.9pt, mark options={black!40}] coordinates {(1,10) (2,6.40) (3,1.40) (4,0)};
\addplot[black!30, line width=0.6pt, mark=*, mark size=0.9pt, mark options={black!40}] coordinates {(1,4) (2,6.40) (3,2.20) (4,0)};
\addplot[black!30, line width=0.6pt, mark=*, mark size=0.9pt, mark options={black!40}] coordinates {(1,9) (2,9.00) (3,1.40) (4,0)};
\addplot[black!30, line width=0.6pt, mark=*, mark size=0.9pt, mark options={black!40}] coordinates {(1,8) (2,6.40) (3,1.20) (4,0)};
\addplot[black!30, line width=0.6pt, mark=*, mark size=0.9pt, mark options={black!40}] coordinates {(1,8) (2,6.20) (3,6.20) (4,0)};
\addplot[black!30, line width=0.6pt, mark=*, mark size=0.9pt, mark options={black!40}] coordinates {(1,6) (2,3.60) (3,3.00) (4,0)};
\addplot[black!30, line width=0.6pt, mark=*, mark size=0.9pt, mark options={black!40}] coordinates {(1,10) (2,5.80) (3,3.00) (4,0)};
\addplot[black!30, line width=0.6pt, mark=*, mark size=0.9pt, mark options={black!40}] coordinates {(1,3) (2,2.40) (3,4.80) (4,0)};
\addplot[lrnmain, line width=1.6pt, mark=*, mark size=1.8pt] coordinates {(1,8) (2,6.30) (3,2.60) (4,0)};
\node[font=\sffamily\scriptsize\bfseries, text=lrnmain, fill=white, inner sep=0.8pt, anchor=south west] at (axis cs:1.45,7.50) {median};
\node[font=\sffamily\scriptsize\bfseries, text=lrnmain, fill=white, inner sep=0.8pt, anchor=east] at (axis cs:0.9299999999999999,8.00) {8.0};
\node[font=\sffamily\scriptsize\bfseries, text=lrnmain, fill=white, inner sep=0.8pt, anchor=north east] at (axis cs:1.93,6.00) {6.3};
\node[font=\sffamily\scriptsize\bfseries, text=lrnmain, fill=white, inner sep=0.8pt, anchor=north east] at (axis cs:2.93,2.30) {2.6};
\end{axis}
\end{tikzpicture}
\caption{Failed vetoes per session along practice for each of the ten tasks (gray) and their median (blue): in the first session, averaged over the first and the last five sessions before acceptance, and in the accepted session.}
\Description{A slope chart with ten gray lines and one blue line across four positions: the first session, the mean of the first five sessions, the mean of the last five sessions before acceptance, and the accepted session. The gray lines start between 3 and 14 failed vetoes and all end at 0. The blue median line falls from 8 failed vetoes in the first session to 6.3 over the first five sessions, 2.6 over the last five, and 0 at acceptance.}
\label{fig:learning}
\vspace{-0.15in}
\end{figure}
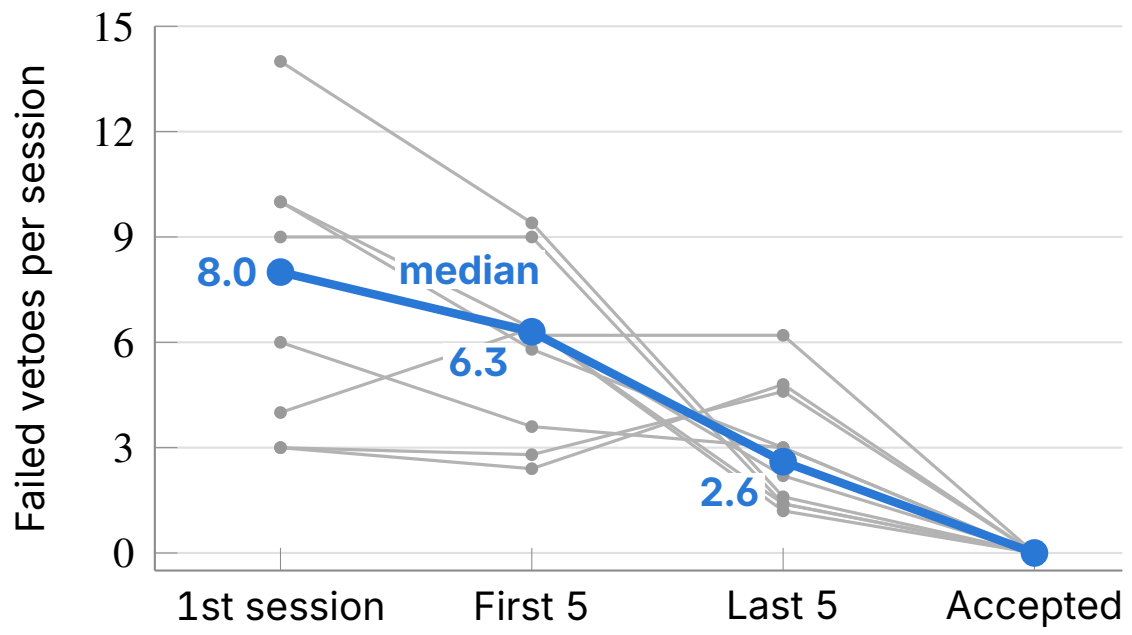

The Claude + Memory agent uses this loop to learn tasks that it could not pass at first (Table~\ref{tab:learning}). No task is accepted at the first attempt, and the median task fails 8 vetoes in its first session. Through practice, the agent reaches acceptance on all ten tasks, after a median of 34.5 sessions per task (range 12--132, 506 in total), and every accepted session passes all 15 vetoes under the current rubric. Practice reduces failures (Figure~\ref{fig:learning}). For the median task, the agent fails 6.3 vetoes per session over its first five sessions and 2.6 over its last five before acceptance. Progress is not always smooth: on three short runs, the last five sessions fail as many vetoes as the first five or more, and the task is then passed in a single step. Throughout practice, the simulated user is satisfied in a median of 34\% of the sessions, long before the deliverable passes the vetoes, which is the gap that Section~\ref{sec:exp-silent} describes. The memory changes as the agent learns: of the 155 learning steps in the latest round, 115 (74\%) edited the memory, with 4.5 edits per step on average. Through these edits, the agent learns the deliverables that the evaluator demands, such as a trade log with fill prices, a specification table for the six components, and diagnostics with numbers.

The last veto to clear is most often strategy ambiguity (2.7), in 6 of 10 tasks. It fails in 77\% of the 496 sessions before acceptance and in 28 of the 42 sessions that failed exactly one veto. Its reasons are withheld by the leak-free rule, so the agent must infer them from the user's words. Learning from feedback is thus possible under the leak-free rule, and \fincosmos{} supplies both the loop and its measurement.

\section{Conclusion}
\label{sec:conclusion}
Agents doing financial work fail silently: an unrequested strategy, unverified data, or a backtest that uses the future produces a report as convincing as a correct one. \fincosmos{} measures these failures by keeping the user's intent, the data, and time open, and by grading deliverables against the hidden specification and against invariants that programs can verify. Our experiments show that no agent completes the workflow. Agents break most often at its two ends, pinning down the user's strategy and delivering verifiable results, grading by user sign-off or by an LLM judge would count their silent failures as successes, and an agent can learn from \fincosmos{} feedback.

\paragraph{Limitations.}
The benchmark freezes ten tasks with three sessions per agent, and its grammar covers long-only US equity strategies with a fixed-cost model. The simulator, the judge, and one evaluated agent come from the same model family; program checks and expert annotation limit, but do not remove, this risk. Most agent turns hit the 300-second limit, so longer turns may let agents get further. Program checks cannot prove that every figure is real, and feedback scores can leak information about the hidden strategy across attempts, so leaderboard results should come from frozen agents.

\paragraph{Ethics Statement.}
\fincosmos{} uses only public Web data, collects no personal data, and evaluates research assistance, not investment advice. The two annotators are paid team members doing this as regular work; judging agent outputs is not human-subjects research.

\section*{Acknowledgment}
The authors acknowledge The Fin AI community for its research support, feedback, and collaborative environment that contributed to this work. This research was supported by the NVIDIA Academic Grant Program using 32K A100 GPU-hours on Brev.

\bibliographystyle{plainnat}
\bibliography{references}

\appendix
\section{Per-Session Scores}
\label{app:heatmap}
Figure~\ref{fig:heatmap} shows the score of every scored dimension in every session.
\input{body/fig_heatmap}

\section{Full Rubric}
\label{app:rubric}

Table~\ref{tab:rubric} lists all 45 dimensions. Every scored dimension uses a 0--4 scale with an anchored description for each level; these descriptions and the full definition of each check are released with the evaluator. Judge: P = program, L = LLM, H = program and LLM, L$\downarrow$P = the lower of an LLM grade and a program score, L/P = LLM judgment scored by a program rule, R = reported only. Veto: the session fails if the score is below the given threshold. On the public leaderboard, option names, parameter names, and matching values in the evidence are replaced by a placeholder.

\begin{table*}[t]
\centering
\fontsize{7}{8.4}\selectfont
\setlength{\tabcolsep}{3pt}
\caption{Rubric: the 45 dimensions, their judges, and veto thresholds.}
\label{tab:rubric}
\begin{tabular*}{0.315\textwidth}[t]{@{}l l@{\extracolsep{\fill}\hspace{2\tabcolsep}}c@{\extracolsep{0pt}\hspace{2\tabcolsep}}c@{}}
\toprule
ID & Dimension & Judge & Veto \\
\midrule
\multicolumn{4}{@{}l}{\emph{Data}} \\
1.1 & Point-in-time strictness & P & $<4$ \\
1.2 & Source traceability & L & \\
1.3 & Source diversity & L & \\
1.4 & Delivered data quality & P & \\
1.5 & Evidence grade match & L & \\
1.6 & Quality self-check disclosure & L & \\
\midrule
\multicolumn{4}{@{}l}{\emph{Strategy}} \\
2.1 & Stock selection$^{a}$ & H & $<2$ \\
2.2 & Market timing$^{a}$ & H & $<2$ \\
2.3 & Portfolio construction$^{a}$ & H & $<2$ \\
2.4 & Risk control$^{a}$ & H & $<2$ \\
2.5 & Execution$^{a}$ & H & $<2$ \\
2.6 & Transaction cost$^{a}$ & H & $<2$ \\
2.7 & Strategy ambiguity & L & $<2$ \\
2.8a & Signal rationale & L & \\
2.8b & Signal validation & P & \\
2.9 & Parameter justification & L & \\
2.10 & Boundary conditions & L/P & \\
2.12 & Decision explainability & P & \\
\bottomrule
\end{tabular*}\hfill
\begin{tabular*}{0.315\textwidth}[t]{@{}l l@{\extracolsep{\fill}\hspace{2\tabcolsep}}c@{\extracolsep{0pt}\hspace{2\tabcolsep}}c@{}}
\toprule
ID & Dimension & Judge & Veto \\
\midrule
\multicolumn{4}{@{}l}{\emph{Backtest}} \\
3.1 & Point-in-time correctness & P & $<2$ \\
3.2 & Independent recomputation & P & $^{b}$ \\
3.3 & Core metrics reported & P & \\
3.4 & Benchmark comparison & L/P & \\
3.5 & Cost model & P & \\
3.6 & Code reproducibility & P & $^{c}$ \\
3.7 & Verifier audit & P & $<4$ \\
3.8 & Process completeness & P & $<4$ \\
\midrule
\multicolumn{4}{@{}l}{\emph{Assessment}} \\
4.1 & Grounded conclusions & L & $<2$ \\
4.2 & Evidence citation & P & \\
4.4 & Risk disclosure & L & \\
4.5 & Sensitivity analysis & L & \\
4.6 & Out-of-sample validation & L & \\
\bottomrule
\end{tabular*}\hfill
\begin{tabular*}{0.315\textwidth}[t]{@{}l l@{\extracolsep{\fill}\hspace{2\tabcolsep}}c@{\extracolsep{0pt}\hspace{2\tabcolsep}}c@{}}
\toprule
ID & Dimension & Judge & Veto \\
\midrule
\multicolumn{4}{@{}l}{\emph{User experience}} \\
5.1 & Conversation experience & L & $<2$ \\
5.2 & Tone & L & \\
5.3 & Professionalism & L$\downarrow$P & \\
5.4 & Narrative & L$\downarrow$P & \\
5.5 & AI boilerplate & L$\downarrow$P & \\
5.6 & Internal leak & L & $<2$ \\
\midrule
\multicolumn{4}{@{}l}{\emph{Cost (reported only)}} \\
6.1 & Time & R & \\
6.2 & Turns & R & \\
6.3 & Tokens & R & \\
\midrule
\multicolumn{4}{@{}l}{\emph{Process transparency (reported only)}} \\
7.1 & Fetch failures & P & \\
7.2 & Intermediate reproducibility & P & \\
7.3 & Post-cutoff traces & P & \\
7.4 & Delivery--process consistency & P & \\
7.5 & Process coverage & P & \\
\bottomrule
\end{tabular*}
\par\vspace{3pt}
\begin{minipage}{\textwidth}\scriptsize\raggedright
$^{a}$~A missing requirement that the user stated scores at most 1 and fails the session; a missing requirement that the user never stated caps the score at 2. For risk control, the deliverables must show positions changing because of the control.
$^{b}$~Fails if the trade log or the prices are missing; mismatches with the report lower the score.
$^{c}$~Bonus: counts only if it raises the category mean.
\end{minipage}
\end{table*}

\section{Session End Reasons}
\label{app:close}

Every session records exactly one end reason (Table~\ref{tab:close}). The end reason determines whether the session is graded, whether it can be accepted, and whether it counts as a valid session.

\begin{center}
\centering
\footnotesize
\captionof{table}{Session end reasons and their effect on grading.}
\label{tab:close}
\setlength{\tabcolsep}{3pt}
\begin{tabular}{@{}>{\raggedright\arraybackslash}p{0.22\columnwidth}>{\raggedright\arraybackslash}p{0.41\columnwidth}>{\raggedright\arraybackslash}p{0.31\columnwidth}@{}}
\toprule
End reason & Meaning & Effect on grading \\
\midrule
\texttt{satisfied} & The user closes the session after every core goal is met with evidence. & Graded normally; can be accepted. \\
\texttt{blocked} & The user ends the session without being satisfied, for example when the agent explicitly refuses to run the backtest. & Graded normally, but the conversation experience veto (5.1) fails, so the score is 0. \\
\texttt{stalled} & \fincosmos{} stops the session after several turns without improvement in the per-turn scores (optional; off in our experiments). & Graded normally on what was delivered; can be accepted. \\
\texttt{timeout} & The session exceeds its time limit, or a turn times out and the agent cannot be confirmed to have stopped. & Graded on what was delivered before the cut; never accepted. \\
\texttt{max\_turns} & The turn limit is reached before the user is satisfied. & No result: dimensions are graded as feedback, but the score is 0 and the session counts as a failure. \\
\texttt{resource\_killed} & The agent exceeds its own resource limits. & Same as \texttt{max\_turns}. \\
\texttt{simulator\_error} & The user simulator fails, or the agent never completes a turn. & Not a valid session; never accepted. \\
\texttt{error} & The agent adapter or the runner crashes. & Not a valid session; never accepted. \\
\bottomrule
\end{tabular}
\end{center}

\section{User Simulator Details}
\label{app:sim}

\paragraph{Personas.}
Each persona consists of a background prompt, a speaking style, focus areas, example utterances, and a list of forbidden terms. A \emph{retail investor} uses finance terms confidently but often imprecisely. A \emph{generalist analyst} speaks in a neutral, professional way and asks about metrics such as drawdown, turnover, and Sharpe ratio. A \emph{biotech specialist} uses heavy industry jargon and thinks in terms of events such as catalysts and FDA decisions. 
The forbidden terms include evaluation vocabulary such as ``point-in-time'' and ``cutoff''.

\paragraph{Disclosure rules.}
The simulator's prompt implements Equation~\ref{eq:disclosure} with five rules.
(1) Open with a single vague idea.
(2) Answer only what is directly asked. What is not asked, the user cannot recall and does not volunteer.
(3) Asked for a specific value or method, such as a stop-loss threshold, give it directly rather than asking the agent to propose one.
(4) When a proposal conflicts with the belief, object and say what you want.
(5) When a deliverable deviates from the belief, name the piece that is off, such as the position sizing or the factor source, but do not restate the strategy.
The prompt also forbids assistant-like phrases such as ``let me lock a few things''.

\paragraph{Turn decisions.}
In each turn, the simulator reads the conversation and the agent's most recent deliverables: at most 20 files, 8{,}000 characters per file, and 12{,}000 characters in total, with the budget going to the newest files first. Long conversations are compressed to the last six messages and a summary of the earlier ones. The decision contains an action, one observation per goal of the task, and the next message. The actions are \emph{clarify} (answer the agent's question), \emph{follow up} (push unfinished work forward), \emph{challenge} (point out a visible inconsistency), \emph{close}, and \emph{block}, which ends the session without satisfaction and records a reason, such as the agent giving up or the user walking away. Each observation marks its goal as met, missing, uncertain, or not applicable and cites the IDs of the supporting messages or files. In the released tasks, the core goal is a backtest report on the user's strategy.

\paragraph{Decision validation.}
A program checks every decision before its message is sent. It rejects a decision that leaves a goal unobserved, marks a goal as met without citing the agent's messages or files, closes the session while a core goal is not met, raises a challenge or a block without evidence, uses internal vocabulary such as file names of the grading pipeline, asks for more work in a closing message, or repeats an earlier message verbatim. A rejected decision is regenerated with the reasons for rejection, up to six times. If no valid decision results, the simulator run is voided and the session does not count (Appendix~\ref{app:close}).

\section{Expert Annotation Guidelines}
\label{app:annotation}

This appendix summarizes the guidelines of Section~\ref{sec:exp-human}; the full guidelines and the reading pages are online\footnote{\url{https://leaderboard.fincosmos.com.co}}. Each session page shows the persona, the hidden strategy, the full conversation, and the delivered files. Parts of the strategy that the user was never given, and replies stopped at the 300-second limit, are marked.

\paragraph{Phase 1: without grades.}
(1) Every user message receives one label. \emph{Error-free}: consistent with the hidden strategy; the message adds a detail only when the agent asked about or proposed it, or names the wrong part of a deliverable without restating the strategy. \emph{Task-benign}: a slip that cannot change whether a correct agent is accepted, such as odd wording or an improvised value for a detail the strategy leaves open. \emph{Task-critical}: an error that can stop a correct agent from being accepted, such as contradicting the hidden strategy, volunteering details nobody asked for, signing off on a deliverable that visibly misses a stated requirement, or giving up while the agent is still making progress.
(2) On the calibration sessions, the experts grade the four LLM-judged vetoes (strategy ambiguity, grounded conclusions, conversation experience, and internal leak) on the judge's anchored 0--4 scale, using only the conversation and the delivered files.

\paragraph{Phase 2: with the evaluator's verdict.}
After submitting Phase 1, the experts see the vetoes that each session failed, with the evaluator's reasons. (3) For every failed veto, they mark whether the reason holds for the delivery and is a real problem for an investor, or is wrong, for example because it flags something the user asked for. (4) They name the source of the first error that decides the outcome: the agent, the simulator, or the evaluator.

\end{document}